\documentclass{article}
\usepackage{ehgan}
\usepackage[utf8]{inputenc} 
\usepackage[T1]{fontenc}    
\usepackage{hyperref}       
\usepackage{url}            
\usepackage{booktabs}       
\usepackage{amsfonts}       
\usepackage{nicefrac}       
\usepackage{microtype}      
\usepackage{lipsum}		
\usepackage{graphicx}
\usepackage{natbib}
\usepackage{doi}
\usepackage{amsfonts}

\title{Efficient Hair Style Transfer with Generative Adversarial Networks}

\author{ {Muhammed Pektas}\\
	University of Ege\\
	Syntonym\\
	Izmir, Turkey \\
	\texttt{https://m-pektas.github.io} \\
	\AND
	{Baris Gecer} \\
	Huawei\\
	London, England \\
	\texttt{https://barisgecer.github.io} 
	\And
	Aybars Ugur \\
	University of Ege \\
	Izmir, Turkey \\
	\texttt{https://avesis.ege.edu.tr/aybars.ugur} \\
}

\begin{document}
\maketitle

\begin{abstract}
Despite the recent success of image generation and style transfer with Generative Adversarial Networks (GANs), hair synthesis and style transfer remain challenging due to the shape and style variability of human hair in in-the-wild conditions. The current state-of-the-art hair synthesis approaches struggle to maintain global composition of the target style and cannot be used in real-time applications due to their high running costs on high-resolution portrait images. Therefore, We propose a novel hairstyle transfer method, called EHGAN, which reduces computational costs to enable real-time processing while improving the transfer of hairstyle with better global structure compared to the other state-of-the-art hair synthesis methods. To achieve this goal, we train an encoder and a low-resolution generator to transfer hairstyle and then, increase the resolution of results with a pre-trained super-resolution model. We utilize Adaptive Instance Normalization (AdaIN) and design our novel Hair Blending Block (HBB) to obtain the best performance of the generator. EHGAN needs around 2.7 times and over 10,000 times less time consumption than the state-of-the-art MichiGAN and LOHO methods respectively while obtaining better photorealism and structural similarity to the desired style than its competitors.
\end{abstract}

\section{Introduction}
Synthesizing hair images and hair style transfer in images has many potential applications such as portraying missing or wanted individuals with different hair styles, previsualization of one's appearance before having a haircut or dyeing, and enhancing appearance for media and editing applications. This technology can even improve visual identity anonymization applications to protect privacy of individuals.

Unlike, most other parts of the human face, its specific geometry, material, and high-frequency details make hair extremely difficult to represent and synthesize.  This complexity comes from the nature of the hair and diverse visual perception factors that the user may intend to edit or preserve. A few key challenges can be identified that are fundamentals to visual quality: 1) complicated hair area boundaries that lead to inadequate high-fidelity hair shape control and seamless background blending, 2) the entanglement between material properties and orientation of hair strands, 3) the correlation between the color palette and strand variations. In addition to the hair complexity, synthesizing high-resolution images, and stabilizing the network training process are other challenging problems~\cite{shaham2021spatially, li2020gan, chai2022any}.

\begin{figure}
\vspace{0.3cm}
\includegraphics[width=\textwidth]{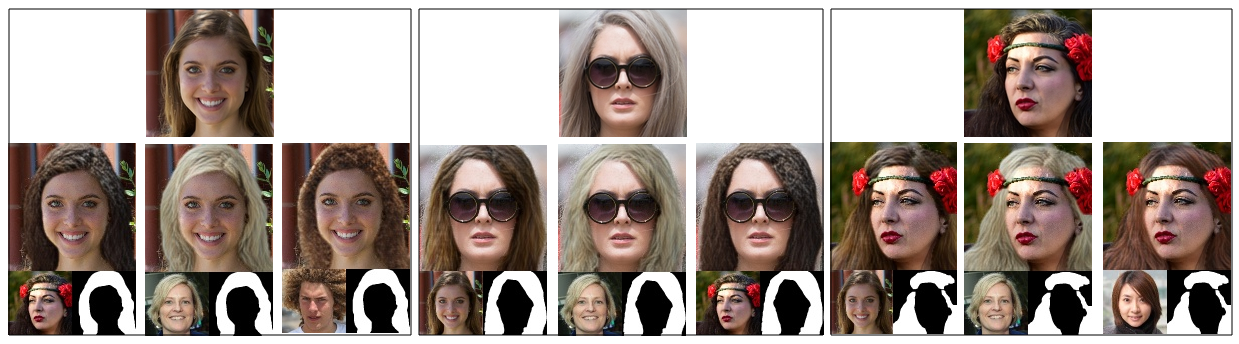}
\caption{Low-resolution (128x128) hairstyle transfer results by our approach. (Top) Input images, (Bottom) Reference image and mask, (Middle) Synthesis by our method.}
\label{fig:intro}
\end{figure}

We identified the most notable two relevant works as MichiGAN~\cite{tan2020michigan}  and LOHO~\cite{saha2021loho}. LOHO decomposes the hair into the perceptual structure, appearance, and style attributes and uses latent space optimization in latent space using the StyleGAN2~\cite{karras2020analyzing} generator while Michigan can control shape, background, structure, and appearance. LOHO has less control ability on hair synthesizing and performs better performance on challenging samples than MichiGAN. Both methods have promising initial results but they are not feasible for real-time applications due to their high computational cost. Moreover these approaches fail to recover global structure of the target hairstyle and synthesize photorealistic hair. 

Our approach aims to improve style transfer and photorealism for hair synthesis compared to these methods while reducing computation time for real-time applications. To achieve this goal, We design a customized low-resolution generator network where we mask its activations with the target hair mask for focusing on only hair region synthesis. We also design hair blending block (HBB) for seamless blending of background and synthesized hair which is followed by a super resolution network. Low-resolution hair style transfer leads to less time consumption while customized generator with HBB leads to performing better hair synthesis and style transfer.

In this paper, we propose efficient hairstyle transfer with a generative adversarial network, called EHGAN. Our method addresses the aforementioned challenges about efficient high-resolution hair synthesis task thanks to our novel framework which performs hairstyle transfer on low-resolution images, seamless blending by HBB module, and increase the resolution of our output using a pre-trained super-resolution network. 
This strategy improves the stability of GAN training for better photorealism, style transfer and time consumption compared to the other methods.

In particular, our main contributions are:

\begin{itemize}

    
 \item We propose a low-resolution style transfer generator followed by super resolution network. In between, the novel Hair Blending Block (HBB) fuse generated hair with the original image for better photo-realistic boundaries. 
  
 \item Qualitative and quantitative experiments show that our method has better hair synthesis results compared to the state-of-the-art.
 
 \item To the best of our knowledge, our method can provide, real-time hair synthesis and transfer  for the first time with 11.6 frame per second efficiency.
  
\end{itemize}

\section{Related Works}

{\bf GAN-based Image Generation.} Thanks to the powerful modeling capacity of deep neural networks, varieties of generative models have been proposed to learn to synthesize images in an unconditional setting. Typical methods include generative adversarial networks (GANs) ~\cite{goodfellow2014generative,mao2017least,metz2016unrolled},  and variational auto-encoders (VAE) ~\cite{doersch2016tutorial,sonderby2016ladder}. Compared to VAE, GANs are demonstrated to be more popular and capable of modeling fine-grained details.  GANs, have been very successful across various computer vision applications such as image-to-image translation ~\cite{isola2017image,wang2018high,zhu2017unpaired,lattas2021avatarme++}, video generation ~\cite{wang2018video,wang2019few,chan2019everybody}, 3D reconstruction~\cite{gecer2021fast}, and data augmentation for discriminative tasks such as object detection ~\cite{li2017perceptual,gecer2018semi,gecer2020synthesizing}. Starting from a random latent code, GANs can generate fake images with the same distribution as natural images in a target domain. The famous work ProgressiveGAN ~\cite{karras2017progressive} first leveraged a progressive generative network structure that can generate very highly realistic facial images, including hair. By further incorporating the latent and noise information into both the shallow and deep layers, StyleGAN and its versions ~\cite{karras2019style,karras2020analyzing,karras2021alias} further improved the generation quality significantly. Particularly on faces, StarGAN's versions can modify multiple attributes. Other notable works, FaceShop ~\cite{portenier2018faceshop}, Deep plastic surgery ~\cite{yang2020deep}, Interactive hair and beard synthesis ~\cite{olszewski2020intuitive} and SC-FEGAN ~\cite{jo2019sc} can modify the images using the strokes or scribbles on the semantic regions.

{\bf Hair Editing and style transfer.} Although some recent work ~\cite{wei2018real,qiu2019two,jo2019sc,lee2020maskgan}, have achieved progress on hair generation conditioned by limited types of inputs, these methods are unfortunately not intuitively controllable and universally applicable. we identified most notable two relevant works as MichiGAN~\cite{tan2020michigan}  and LOHO~\cite{saha2021loho}. MichiGAN proposed a conditional synthesis GAN that allows controlled manipulation of hair. MichiGAN disentangles hair into four attributes by designing deliberate mechanisms and representations and produces SOTA results for hair appearance change. LOHO decomposes the hair into perceptual structure, appearance, and style attributes and uses latent space optimization~\cite{karras2020analyzing,gecer2021ostec} to infill missing hair structure details in latent space using the StyleGAN2 generator. While Michigan can control shape, background, structure, and appearance. LOHO has less control ability on hair synthesizing and performs better performance on challenging samples than MichiGAN. Although, EHGAN has less control ability than MichiGAN, it requires less processing time than both methods while obtaining better overall hair style transfer results. Except for these methods, there are several notable works about hairstyle transfer and synthesizing~\cite{fan2021hsegan,zhu2021barbershop}.

\section{Method}

\subsection{Overview}

Given an input portrait image $I_{inp}$ and reference portrait image $I_{ref}$, our EHGAN aims at performing conditional hairstyle transfer to the hair region according to reference input, while  keeping the background $I_{bg}$ unchanged. The output is denoted as $I_{out}$.

\begin{figure*}
   \includegraphics[width=\textwidth]{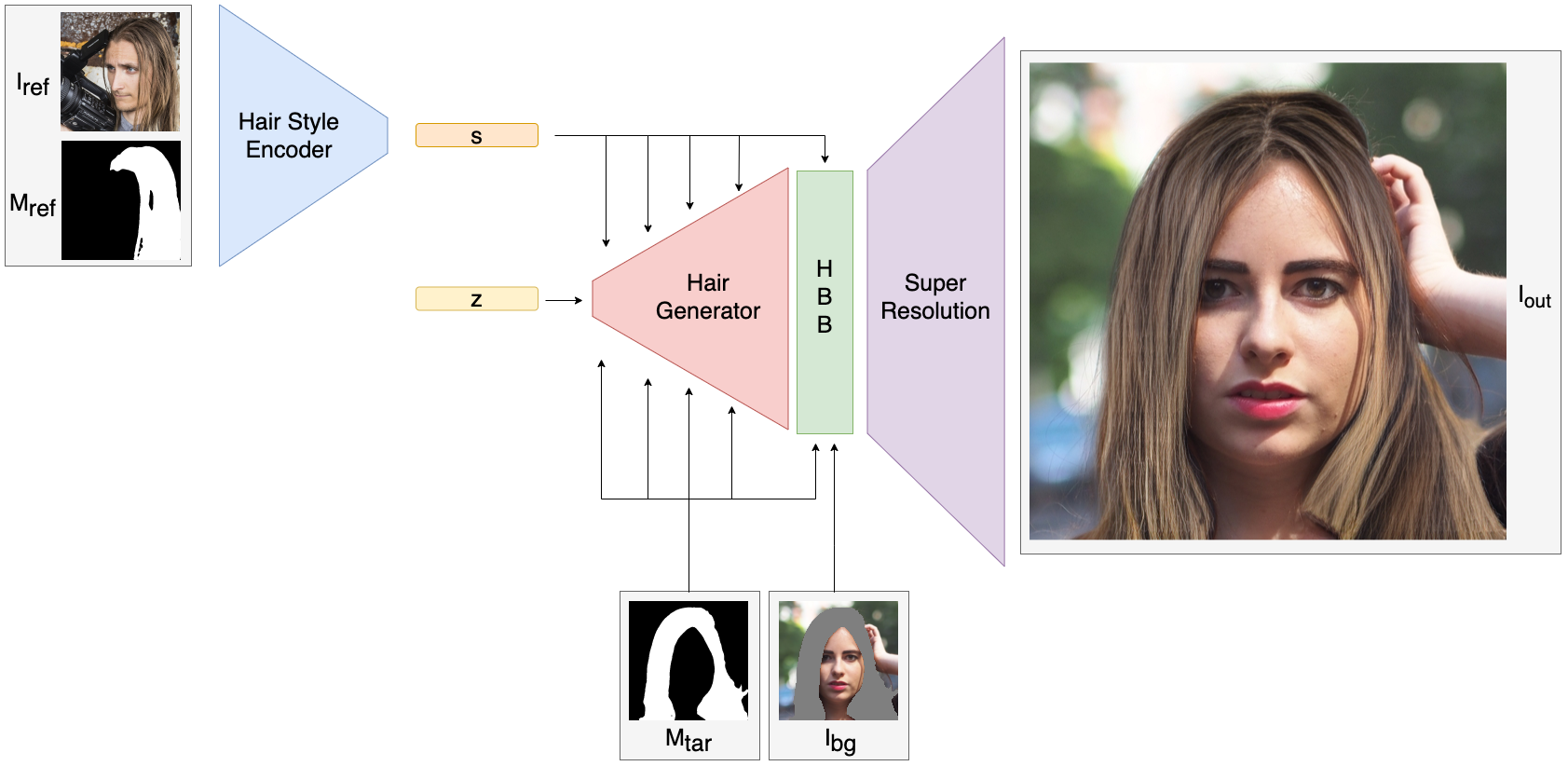}
   \caption{ The hairstyle encoder $E$ estimates the reference hairstyle vector $s$ by taking reference hair $H_{ref}$ as an input image that is obtained by multiplying the reference image $I_{ref}$ and reference hair mask $M_{ref}$. The hairstyle vector $s$, target hair mask $M_{tar}$,  input background $I_{bg}$ and noise vector  ${z}$ feed to the generator $G$ the generator generates a fake hair image $H_{fake}$. Finally, low-resolution hair transferred image resolution was increased by a pre-trained super-resolution network to obtain the final image $I_{out}$. }
\label{fig:EHGAN}
\end{figure*}

As illustrated in Figure \ref{fig:EHGAN}, the proposed pipeline consists of generation, hairstyle encoding, and super-resolution parts. Super-resolution is used only for the inference stage. In the training stage, the hairstyle encoding part is responsible for the estimation of the hairstyle representation of the reference image. The generation part synthesizes hair style transferred image. In our method, we estimate reference hairstyle representation $s$ by style encoder network $E$. Style encoder network needs to reference hair region $H_{ref}$. Generator $G$, needs this style representation, random sampled noise $z$, input background $I_{bg}$ and target hair shape mask $M_{t}$. Finally, we increase synthesized image resolution by pre-trained super-resolution network $SR$. We can formulate our hair synthesis process as:

\begin{equation}
I_{out} = SR(G(z,M_{t},I_{bg},E(H_{ref})))
\end{equation}

\subsection{Hair Style Encoding}

Hair appearance describes the globally-consistent color style that is invariant to a specific shape or structure. It includes multiple factors such as the intrinsic albedo color, environment-related shading variations, and even the granularity of wisp styles. Due to its global consistency, its appearance can be compactly represented. Given the complexity of natural hairstyles, instead of generating the appearance features from scratch, we address hair appearance as a style transfer problem, which extracts style from the reference image and transfers it to the target.

First, we multiply reference image and mask each other to extract reference hair region $H_{ref}$ . Than, we resize $H_{ref}$ from current resolution to our resolution size for style encoding network. Style encoding network consist of residual blocks and it estimates hair style representation $s = E(H_{ref})$. $E(.)$ corresponds the style encoding estimation and $s$ corresponds to style representation vector $s \in \mathbb{R}^{Nx512}$ of $H_{ref}$.

\subsection{Image Synthesis}

For providing an efficient hairstyle transfer framework, we transfer hairstyles to the low-resolution image. So, we used a low-resolution hair style encoder, generator, and discriminator networks. EHGAN contains two main modules in training: a hair synthesis generator $G$ and a hairstyle encoder $E$. The hair synthesis generator has two types of blocks that are Adaptive Instance Normalization Resnet Block (AdaINResBlk)  and Hair Blending Block (HBB).

\subsubsection{Low-resolution image synthesis}

Before the synthesizing to hairstyle transferred image, we resize the input image background $I_{bg}$ to our resolution size. Our generator $G$ synthesize the transferred image corresponding to the hairstyle of reference hairstyle $s_{ref}$, which is provided by style encoding network $E$. The generator is a decoder architecture that is modified for hair synthesis tasks. It consisted of residual blocks and a hair blending block (HBB). We believe that inserting style information only at the beginning of a network is not a good architecture choice. Recent architectures ~\cite{karras2019style,choi2020stargan} have demonstrated
that higher quality results can be obtained if style information is injected as normalization parameters in multiple layers in the network, e.g. using adaptive instance normalization (AdaIN) ~\cite{huang2017arbitrary}. Therefore, We use AdaIN in each residual block to inject $s_{ref}$ into $G$. We called AdaINResBlk ~\cite{choi2020stargan,huang2018multimodal} this combination of residual block and AdaIN.
\begin{figure}
\begin{center}
\includegraphics[width=0.5\textwidth]{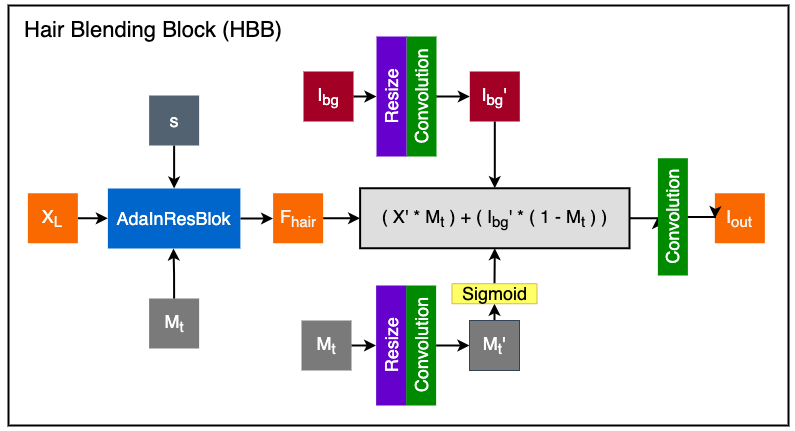}
\caption{Hair blending block (HBB). It combines synthesized hair area and background seamlessly.}
\label{fig:HBB}
\end{center}
\end{figure}

{\bf AdaINResBlok} requires three inputs,  layer activation $X_{L}$, a target hair mask $M_{t}$ to focus only hair area for hair generation, and a reference hairstyle $s_{ref}$ to transfer reference hairstyle information. For the first layer, $X_{L}$ is a gaussian noise vector $z \in \mathbb{R}^{Nx512}$. We mask the activation of AdaINResBlock, multiplying with the target hair mask. The generator has a better focus on target hair shape.

{ \bf Hair Blending Block (HBB)} designed to perform seamless hair blending. The output of the last AdaINResBlk contains only synthesized hair area.  We need to blend this synthesized hair and input background regions. Therefore,  HBB requires the background of input image $bg_{inp}$, hairstyle vector $s$ to transfer reference hairstyle, and synthesized hair area $F_{hair}$. This block contains AdaINResBlock and extra operations. We feed input background and target hair mask into HBB to obtain better seamless blending performance. We add learnable convolutional layers editing background and hair mask image. HBB performs hair blending in only one layer differentially. Also, It is suitable with generators that use all capacities to generate a hair area of the image. Finally, we extend the HBB with a final
3x3 convolution layer, with a step size of 1 and 1 pixel
padding, to refine the final output and reduce noise. HBB is shown in Figure \ref{fig:HBB}.

\subsubsection{Super-resolution}
In our efficient hairstyle transfer setting, we use a low-resolution style encoder and generator networks. For this reason, we need super-resolution layers to generate high-quality hair synthesis results. Instead of the training layers that perform the super-resolution task, we use state-of-the-art pre-trained GFPGAN~\cite{wang2021gfpgan}. Other methods~\cite{tan2020michigan,saha2021loho} synthesize results that have 512x512 resolution. But, our trained generator and encoder models take 128x128 input. So, we need to increase 4x resolution of our results to obtain comparable results with their results.


\subsection{Objectives}

{\bf Adversarial loss:} A two-scale PatchGAN \cite{isola2017image, wang2018high} discriminator D is used to match distributions between synthesized hair results and real hair to enforce
natural visual quality and boost local details. Loss can formulate as:

\begin{equation}
    L_{adv}= \left[ log (D(I)) \right] +\left[ log(1-D(G(z,s,I_{bg},M_{inp} ) \right]
\end{equation}


{\bf Hair loss:} We also the penalize both photometric and feature differences between reference and predicted hair region of image by hair loss. We utilize from  pre-trained VGG19 model \cite{simonyan2014very} for extractiong feature. VGG19 is widely used in previous image generation problems ~\cite{wang2018esrgan,saha2021loho}.

\begin{equation}
L_{hair}=  \left\| Vgg(H_{ref})-Vgg(H_{fake}) \right\|_{1}*\lambda_{hairperceptual} + \left\| H_{ref} - H_{fake} \right\|_{1} * \lambda_{hairpixel}
\end{equation}

{\bf Pixel Loss:}
While the optimized natural visual quality and high-level features, colors, and lighting conditions are optimized based
on pixel value difference directly. While this cost function
is relatively primitive, it is sufficient to optimize lighting and colors. Furthermore, pixel loss supports natural visuality and high-level features. It formulates in pixel-level as:

\begin{equation}
    L_{pix}=\left\| I_{inp} - I_{fake} \right\|_{1}
\end{equation}

{\bf Style reconstruction:} In order to enforce the generator $G$ to utilize the style code $s$ when generating the hair $I_{out}$,
we employ a style reconstruction loss as:

\begin{equation}
    L_{style}=   \left\| E(H_{ref}) - E(H_{fake})) \right\|_{1}
\end{equation}

Thanks to this loss, Encoder learns a mapping from a hair to its hairstyle vector and the generator learns to generate the same hair images for the same hairstyles. Thus, at test time, our learned encoder $E$ allows $G$ to transform an input hair, reflecting the style of a reference hair.

In summary, the overall training objective can be formulated as:
\begin{equation}
    min_{G} max_{D} (\lambda_{pix}L_{pix} + \lambda_{hair}L_{hair} + \lambda_{adv}L_{adv} +\lambda_{style}L_{style} ) 
\end{equation}

\subsection{Training Strategy}

Considering that we do not have the ground truth images when the conditional input comes from different sources, training the proposed EHGAN is essentially an unsupervised learning problem. However, most existing unsupervised learning algorithms \cite{liu2019few,zhu2017unpaired, huang2018multimodal} are usually performing worse than supervised learning regarding the visual quality of the generated results. Motivated by \cite{he2018deep,tan2020michigan}  which faces a similar problem on image colorization and hair synthesis, we propose to train our EHGAN in a pseudo-supervised way. Specifically, we feed the conditional input extracted from the same source image into EHGAN and enforce it to recover the original image with explicit supervision during training, and generalize it to arbitrary reference hairstyle in inference.

\section{Experiments and Results}
\subsection{Implementation Details}

{\bf Datasets:} We use the Flickr-Faces-HQ dataset (FFHQ) \cite{karras2019style} that contains 70000 high-quality images of human faces. Flickr-Faces-HQ has significant variation in terms of ethnicity, age, and hairstyle patterns. The FFHQ dataset is split into the training set of 56000 images and the testing set of 14000 images, with 512×512 resolution. We resize all datasets 128x128 resolution to our efficient model training.
\\
{\bf Training Parameters:} We train style encoder network and the generator network jointly. The Adam optimizer \cite{kingma2014adam} is used with a batch size of 8 and the total epoch number of 55. The learning rates for the generator and discriminator are set to 0.0001. All loss weights,  $\lambda_{pix}$ , $\lambda_{perc}$, $\lambda_{adv}$, $\lambda_{style}$   are simply set to 1.
\\
{\bf Comparisons to Other Methods:}
We evaluate our method by comparing the state-of-the-art methods, LOHO and MichiGAN. The authors of LOHO and MichiGAN provide public implementation, which we used in our comparison. LOHO's authors used public models for masking~\cite{Gong2019Graphonomy} and inpainting~\cite{yu2018free}. In the comparison, all hyperparameters and configuration options were kept at their default values for both methods. 
 
\subsection{Qualitative Experiments}
\begin{figure}
\centering
  \includegraphics[width=0.74\linewidth]{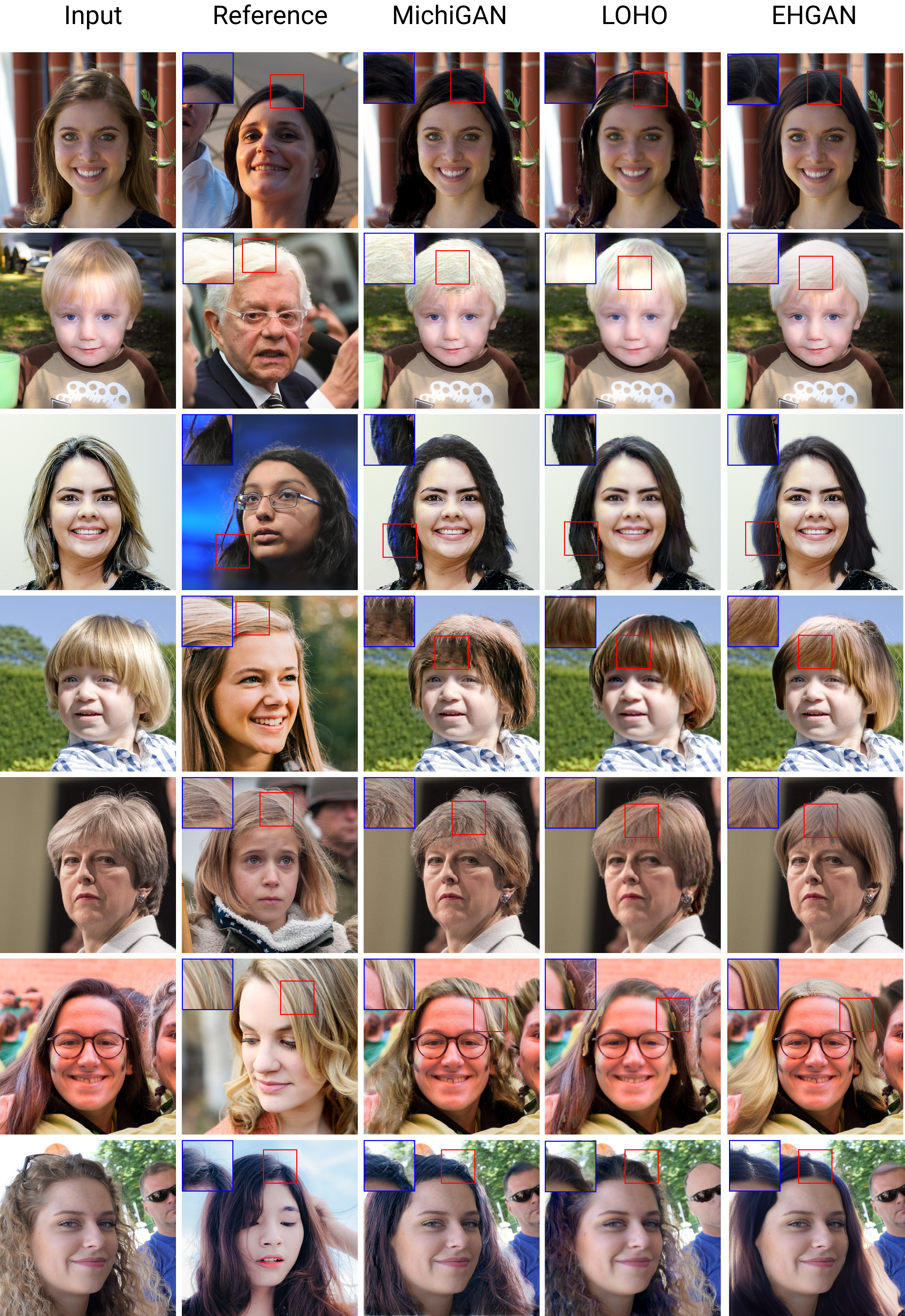}
  \caption{Style transfer comparison of LOHO, MichiGAN and EHGAN qualitatively. Red and blue boxes shows zoomed version of a patch.}
\label{fig:styletransfer}
\end{figure}

Qualitative experiments in Figure \ref{fig:styletransfer} show that, our method has superiority over other methods in terms of hair style transfer and photorealism. Particularly, our method has provided better hair part transfer from reference hair style in $1^{st}$, $6^{th}$ and $7^{th}$ rows, fabric in $2^{nd}$ and $5^{th}$ rows, and generally less artefacts in $3^{rd}$ and $4^{th}$ rows.

\begin{figure}
\centering
  \includegraphics[width=0.82\textwidth]{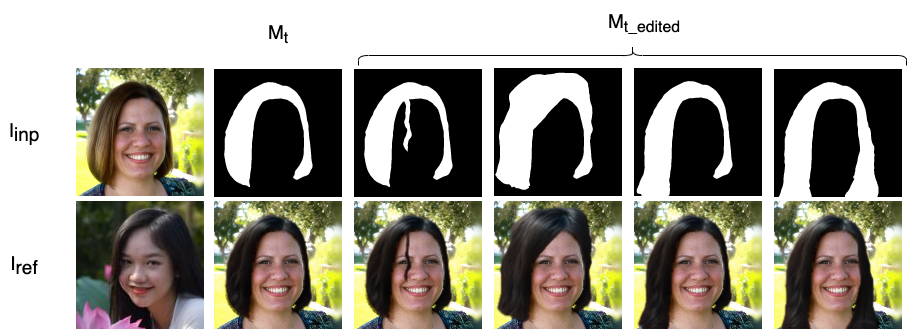}
  \caption{Multi-attribute editing results of our method.}
 \label{fig:mutliatttrasnfer}
\end{figure}

In addition, our method is capable of editing attributes of in-the-wild portrait images. In this setting, an image is selected and then an attribute is edited individually by providing reference images. We can transfer reference image style and appearance to input portrait images and can edit hair shape. We don't use any inpainting module like other methods. Therefore, our shape editing ability is limited as the target hair shape should cover to original hair area for best hair shape editing. But further works, our method can extend easily with inpainting models. Our method changes only the target hair area while don't change other areas such as background, face, etc. Our and other methods' qualitative results are shown in Figure \ref{fig:mutliatttrasnfer}.

\subsection{Quantitative Experiments}

For quantitative experiments, we process a random subset of FFHQ~\cite{karras2017progressive} dataset with 5000 images by related methods and evaluate the synthesized images by the metrics. We calculate visual quality using FID\cite{heusel2017gans}, PSNR, and SSIM metrics. According to FID, EHGAN has the best results with 7.384 but LOHO has the highest performance on PSNR and SSIM. The quantitative evaluation comparison is shown in Table \ref{table:fid}. 





\begin{table}[t]
\begin{minipage}{\textwidth}
\begin{minipage}[b]{0.47\textwidth}
\centering

\begin{tabular}{llll}
\hline
Methods & FID & PSNR & SSIM\\

\hline

MichiGAN & 10.697 & 33.44  & 0.7910   \\
LOHO & 8.419 &  32.2 & {\bf 0.93}  \\
{\bf EHGAN} & {\bf7.384} & {\bf34.767}  & 0.8177  \\

\hline
\end{tabular}
\vspace{0.4cm}
\caption{{ \bf FID, PSNR and SSIM} for different methods. We use 5000 images randomly sampled from the
testing set of FFHQ.  Lower ($\downarrow$) PSNR and FID are better and higher ($\uparrow$) SSIM is better.}
\label{table:fid}

\end{minipage}
\hfill
\begin{minipage}[b]{0.47\textwidth}

\label{table:time}
\begin{tabular}{lll}
\hline
Methods & Runtime (s)& FPS\\

\hline

MichiGAN & 0.2329 & 4.29\\
LOHO & 1447.9  & 0.0007\\
{\bf EHGAN} & { \bf 0.0862} & { \bf 11.6 }\\

\hline
\end{tabular}
\vspace{0.4cm}
\caption{{ \bf Second per image (sec)} for different methods. Lower ($\downarrow$) sec is better. }
\label{tab:timeconsuption}

\vspace{1.23cm}
\end{minipage}

\end{minipage}
\end{table}

\subsubsection{Time Consumption}
In terms of time consumption, our method has state-of-the-art performance. We compare methods by runtime performance as second which is the sum of inference and post-processing time. We don't involve preprocessing time because preprocessing step can change depending on the different applications.  Our method needs 0.0862 seconds for a hairstyle transfer but LOHO and MichiGAN need 1447.9 and 0.2329 respectively. According to Table \ref{tab:timeconsuption}, our method has almost 2.7x and 16796x less time-consuming than other methods on NVIDIA RTX 2080 GPU.

\section{Discussion}

In this work, we propose EHGAN which performs hair style transfer. We design HBB to obtain seamless boundaries between background and synthesized hair while the generator focusing the target hair region. In EHGAN, We perform hairstyle transfer on low-resolution images and then increase resolution 4x by a pre-trained super-resolution model. Thanks to our all framework and designed generator with HBB, EHGAN needs less time consumption while synthesizing hair that is competitive with state-of-the-art results. 

We compare our method in terms of visual quality and runtime performance. We demonstrate the visual quality of our results quantitatively by FID, PSNR, and SSIM metrics in Table \ref{table:fid} and qualitatively in Figure \ref{fig:styletransfer}. According to our comparison, EHGAN performs competitive hair style transfer performance while needing 2.7x and 16796x less time consumption than MichiGAN and LOHO respectively.

Currently,  state-of-the-art hair synthesis methods can support only user-edited hair synthesis applications. We believe that more fast and robust hair style transfer helps to discover more diverse hair synthesis applications. So, EHGAN demonstrates that hair synthesis can be a more fast process while obtaining state-of-the-art results. In terms of this, We suppose that EHGAN can unlock the potentials for fast and robust hair style transfer methods.

\bibliographystyle{unsrtnat}
\bibliography{ehgan}

\end{document}